\pdfoutput=1

\documentclass[11pt]{article}

\usepackage[]{acl}

\usepackage{times}
\usepackage{latexsym}

\usepackage[T1]{fontenc}

\usepackage[utf8]{inputenc}

\usepackage{microtype}

\usepackage{mdframed}

\title{Language Models Understand Us, Poorly}

\author{Jared Moore \\
  University of Washington School of Computer Science \\
  \texttt{jared@jaredmoore.org}}

\usepackage{lipsum}
\begin{document}
\maketitle
\begin{abstract}

Some claim language models understand us. Others won't hear it.
To clarify, I investigate three views of human language \textit{understanding}: \textit{as-mapping}, \textit{as-reliability} and \textit{as-representation} (\S \ref{sec:views}). I argue that while behavioral reliability is necessary for understanding, internal representations are sufficient; they climb the right hill (\S \ref{sec:hill}). I review state-of-the-art language and multi-modal models: they are pragmatically challenged by under-specification of form (\S \ref{sec:under-spec}). I question the \textit{Scaling Paradigm}: limits on resources may prohibit scaled-up models from approaching understanding (\S \ref{sec:scaling}). 
Last, I describe how \textit{as-representation} advances a science of understanding. We need work which probes model internals, adds more of human language, and measures what models can learn (\S \ref{sec:discussion}). 


\end{abstract}

\section{Introduction}

A theme of EMNLP this year is "unresolved issues in NLP." Hence I consider what it means to understand human language, whether current language models understand and whether future models will.

Recent large language models have achieved impressive results on benchmark tasks \citep{thoppilan_lamda_2022, brown_language_2020}.
These results challenge ordained wisdom on the representations necessary for language production. We've seen improved results from multi-modal models \citep{saharia_photorealistic_2022, ramesh_hierarchical_2022, ramesh_zero-shot_2021, shuster_image-chat_2020, radford_robust_2022, borsos_AudioLM_2022}, what some call foundation models \citep{bommasani_opportunities_2021}. Some models even run images, text, and games \cite{reed_generalist_2022}. \citet{michael_what_2022} identify language understanding and scaling as pertinent and much debated questions in NLP.

So what's next? I identify three views on language understanding (\S \ref{sec:views}): \textit{understanding-as-mapping}, \textit{ understanding-as-reliability}, and \textit{understanding-as-representation}.
Through examples of recent limitations of language models (\S \ref{sec:under-spec}), I argue for \textit{understanding-as-representation} because it climbs the right hill (\S \ref{sec:hill}). In particular, I question the assumption that scaling current models is computationally feasible to lead to human-like understanding (\S \ref{sec:scaling}). Because of the large gap between human and model understanding, I think it is generally misapplied to say that models "understand" (\S \ref{sec:imply}). Better applied are examples of promising work on understanding (\S \ref{sec:future}).



\section{Views on Understanding}\label{sec:views}

Some argue that there is a strict barrier which separates human from machine understanding \citep{bender_climbing_2020, searle_minds_1980}.
\textbf{Understanding-as-mapping} puts \textit{understanding in terms of an absolute mapping between form and meaning}. Here, meaning comes from what a series of forms describes. Those forms can be composed in a variety of ways to yield different, legible meanings.\footnote{\citet{goldberg_compositionality_2015} reviews compositionality.}
Often, those with this view imply humans have special access to meaning. 



Others argue that we ought be rid of the distinction between human and machine understanding. They imply models will close the gap soon enough \citep{ manning_human_2022, aguera_y_arcas_large_2022, kurzweil_singularity_2005, turing_computing_1950}. \textbf{Understanding-as-reliability} puts \textit{understanding as a question of
reliable communication}: can one agent expect another agent to respond to
stimuli in a certain way?\footnote{\citet{michael_dissect_2020} names this the behaviorist view.}
This view assumes that scaling alone will lead to an agent capable of human-like language; system internals don't matter. For example, in the most extreme case we can imagine a very large look-up table with state (cf. \citealt{russell_artificial_2021}): a mapping from every input sequence to a sensible output sequence.

In this paper, I put \textit{understanding in terms of internal, dynamical representation}: when prompted with a stimulus, does an agent reproduce an internal representation similar enough to that intended?
Call this \textbf{understanding-as-representation}.
Many have proposed related theories \citep{shanahan_abstraction_2022,barsalou_grounded_2008, hofstadter_surfaces_2013, jackendoff_users_2012, grice_studies_1989}. 
In this view, if someone unthinkingly blurts out the correct answer to a question, they would not have understood. While
a thermostat reproduces a certain representation given a temperature this representation is not similar to a person's.
Some have said that models appear not to understand
because their interrogators fail to present stimuli in a model-understandable way (\citealt{michael_dissect_2020} summarizes).
Exactly: I am concerned with human language understanding--not any possible form of understanding.

To advance a science of understanding, I argue that \textit{as-reliability} is necessary, \textit{as-representation} is sufficient, and \textit{as-mapping} is neither.

I reject the premise of \textit{as-mapping} that the way we use words is separate from our meanings.
While current work in NLP poorly approximates shared intentionality\footnote{The meaning to which \citet{bender_climbing_2020} says models have no access.} I disagree that this is the only route to meaning.\footnote{Millikan offers an account where inner representations exist but are not shared \cite{millikan_beyond_2017}.} We could imagine a very large look-up table.
There is no boundary between what is and what is not a language.\footnote{\citet{bender_climbing_2020} permit meaning in models which ground linguistic form on images.}




I accept \textit{as-reliability} in theory. Enough data and parameters should yield a language-performant agent indistinguishably similar to a human tested on byte streams passed along a wire. Similarly, \citet{potts_could_2022} argues that a self-supervised foundation model could do so.
Still, I am skeptical of what I call the \textit{Scaling Paradigm}, that scale alone is a realistic approach.

I think that hill climbing works but we're climbing the wrong hill.

\section{Climbing the Right Hill}\label{sec:hill}

\textit{As-representation} and \textit{as-reliability} are compatible: we may care about \textit{representation} but more easily look for \textit{reliability}. I argue that input-output behavioral tests are necessary but may not be sufficient to attribute understanding--we may need to look inside.\footnote{Compare \citet{churchland_could_1990}.}

Nonetheless, Alisha, when messaging with Bowen,
has no need to look inside Bowen's head to verify that he understood the following exchange:

\begin{quote}
A: I'm unhappy.\\
B: Why aren't you happy?
\end{quote}

\noindent Our human bias is to assume that other agents understand until evidence proves otherwise \citep{weizenbaum_computer_1976}. This is pragmatic; until recently humans did not encounter non-human agents who could respond somewhat reliably.
\textit{Humans assume a similarity of representation}, that others have the same inductive biases.

\textit{We can't make that assumption with our models.} We can't assume that a chat-bot has a bias to coo over babies (cf. \citealt{hrdy_mothers_2009}). This is why Turing's (\citeyear{turing_intelligent_1948}) test doesn't work--the smoke and mirror programs which won the Loebner prize unintentionally parody input-output tests \citep{minsky_annual_1995}. Reliability, while useful, alone does not advance a science of understanding. \textit{As-reliability} does not tell us which biases induce understanding. It is not causal.




Granted, humans' internal representations are difficult to measure, may change at each point of access,
and in AI we've historically leaned too heavily on certain putative representations. \citet{sutton_bitter_2019} calls this a "bitter lesson."


So why talk of representation? I agree with the "bitter lesson" but I also know that there is no such thing as free lunch; human language occupies a small manifold in the space of possible functions. I don't argue to replicate natural functions but rather to be honest about human strengths lest we wander off into fruitless regions of state space.
To do logic, at some internal level a system is going to have to appear to use the parts of logic.

Advancing \textit{as-representation} does not mean we know what representations underlie human language nor that we must use certain ones.



Advancing \textit{as-representation} does mean that we pay attention to the constraints on human language usage (\S \ref{sec:under-spec}). We should use those to guide our benchmark tests for reliability. We should not get lost in our proxies, especially what the \textit{Scaling Paradigm} assumes (\S \ref{sec:scaling}).





\section{Under-specification of Meaning}\label{sec:under-spec}


Language is dynamic (e.g. has a history), intersubjective (multi-agent), grounded in a large number of modalities (senses), collectively intentional (in a cultural context), and more. Present models have little, if any, data on these aspects of language.

As \citet{bisk_experience_2020} make clear in their "world scopes," the majority of work in NLP attempts to learn language from internet text alone.
I agree that models have fewer data of the world than humans \citep{bender_climbing_2020}. 
What our models see under-specify our meanings.

Recent work has looked into such dissimilarities.
\citet{mccoy_how_2021} note that while language models use novel
constructions, they copy from training data at a high rate.
\citet{mccoy_right_2019} and \citet{branco_shortcutted_2021} identify how models use heuristics and short-cuts contrary to the human meaning of prompts.
\citet{shaham_scrolls_2022} show that a current model will not remember a game after a long-enough context window.

Work has just begun to show limitations of multi-modal models. \citet{thrush_winoground_2022} test for compositionality over images and text: current models perform at chance. \citet{marcus_very_2022} and \citet{conwell_testing_2022} show many similar compositional issues for DALLE-2 in comparison to humans. \citet{lake_word_2021} note the implausibility of present multi-modal models: e.g. they cannot describe internal desires or change beliefs.




Thus my claim is not that models can't learn meaning. My claim is that for models to approach human meaning they will require data on aspects of language the field has only begun to investigate. Consider two examples of under-specification:





\textbf{Under-specification of physics}.
A model over text and static images would perform poorly on a query such as, \textit{"Can you remove this block without causing the tower to fall?"} paired with an image where a finger points at a block that could obviously be removed (or obviously not).



\textbf{Under-specification of time}. In Western contexts respondents associate earlier to the left and later to the right.
This is not specified in language and is mutable (\citealt{casasanto_mirror_2014}, and for the example).
Thus I expect models which have no notion of time
to perform poorly on tests of temporal bias such as those in Fig. \ref{fig:temporal}.


\begin{figure}
Prompt: are each of these events temporally earlier (A), present (B), or later (C)?

\begin{minipage}{.22\textwidth}
\begin{tabular}{|l|l|l|}
\hline
\multicolumn{3}{|l|}{\textit{one day after}} \\

A  &  B  & \textbf{(C)} \\ \hline
\end{tabular}
\begin{tabular}{|l|l|l|}
\hline
\multicolumn{3}{|l|}{\textit{retfa yad eno}}\\

\textbf{(C)} & B & A \\ \hline
\end{tabular}
\end{minipage}
\begin{minipage}{.22\textwidth}
\begin{tabular}{|l|l|l|}
\hline
\multicolumn{3}{|l|}{\textit{a day before}} \\

\textbf{(A)} & B & C \\ \hline
\end{tabular}
\begin{tabular}{|l|l|l|}
\hline
\multicolumn{3}{|l|}{\textit{erofeb yad a}} \\

C  &  B  & \textbf{(A)} \\ \hline
\end{tabular}
\end{minipage}
\caption{Stimuli to probe temporal biases. Answers \textbf{(bolded)}. In human subjects response times are shorter, once trained, for pairings shown and longer when the order of the answers is reversed. A model with similar temporal bias should assign a higher probability to the correct answer in the intuitive ordering (shown).}
\label{fig:temporal}
\end{figure}



\section{Challenges to the \textit{Scaling Paradigm}}\label{sec:scaling}

But what of more data? The \textit{Scaling Paradigm} tells us that scale alone--more parameters, more data, more modalities--will be enough to approximate human language understanding. Exponential increases in parameters or training data have yielded linear increases in performance \citep{chowdhery_palm_2022, kaplan_scaling_2020}.


Still, text doesn't seem like enough. \citet{merrill_provable_2021} argue that there aren't enough examples to learn meaning from form in languages of assertions. Even \citet[pg. 48]{chowdhery_palm_2022} admit to be running out of clean data for exponentially-bigger models. Furthermore, by the age of five, the average American child has heard between ten and fifty million words \citep{sperry_reexamining_2019}.
A state-of-the-art model sees from 10 to 100k more words than a kid.\footnote{GPT-3 uses .3 trillion tokens \citep{brown_language_2020}, LaMDA 2.8 trillion \citep{thoppilan_lamda_2022}, and PaLM .7 \citep{chowdhery_palm_2022}.}  For some (e.g.  \citealt{linzen_how_2020}), the argument stops there: our language models must not be learning the correct functions because they require more data to generalize. 

Nonetheless, humans have plenty of data to compare language use besides the words they hear \citep{tomasello_constructing_2003,lakoff_metaphors_1980}.
For example, \citet{smith_developing_2018} show that children at first need items to be visually centered to learn them.
So the claim that models learn the wrong function may only apply when limited to text.

What of more modalities, then?
To extrapolate on the figures of PaLM \citep[\S 13]{chowdhery_palm_2022}, we only need to scale a model to about $2^{12}$ billion parameters and train it for $2.59 \times 10^{26}$ flops (and increase training data) for perfect performance on a variety of English NLP tasks. For perspective, \citet{heim_estimating_2022} estimates the cost of PaLM at 10 million USD  which, at the 100x projected, is 1 billion. These figures seem infeasible.

Furthermore, the tests in PaLM do not capture the limits I mentioned for human-like understanding (\S \ref{sec:under-spec}); we would need to add image \citep{ramesh_hierarchical_2022} and game \cite{reed_generalist_2022} networks as well.
I don't know exactly what effect these added modalities will have except that they will increase the exponent on scale. I am wary that the long tail of returns embraced by the \textit{Scaling Paradigm} chases an exponential.\footnote{While some note the exponential scales of large models \citep{thompson_deep_2021, john_ai_2022} they may not account for counter-measures \citep{patterson_carbon_2022}.}
The primacy of scale implies an exponentially diminishing longtail of capability.
How soon until our models become planetary in size, as \citet{bostrom_superintelligence_2014} portends?


I thus recast the debate, not as what can theoretically be learned from data (as the \textit{Scaling Paradigm} trumpets),
but as the computational efficiency of different learning approaches. \textit{To asymptotically approach an approximation of human language understanding, how many parameters, data, and modalities will our models need?}












These concerns about scale make me hesitant to suggest that efforts will soon close the gap between human and machine understanding (contra \textit{as-reliability}) even as I agree that they will narrow it (contra \textit{as-mapping}).

\section{Discussion}\label{sec:discussion}

\subsection{Sorta Understands != Understands}\label{sec:imply}




To admit models may one day understand like us and to claim they now understand in limited domains, need not lead us to give understanding over to machines far to one end of the spectrum. These models understand parts of language---just not in the same way as humans understand each other. They are not biased in the same way humans are. Machines will make unknown mistakes unless we can interrogate their representations.


Moreover, attributing "understanding" is consequential. 
Some argue for granting moral rights to robots \citep{gordon_moral_2021}. Page describes those decrying AI as "speciesist" \citep{tegmark_life_2017}. 
\citet{bryson_robots_2010} argues to the contrary.
But note that understanding
and moral status aren't the same. I urge caution over assuming the distinction between human and machine will disappear anytime soon.

We should not let the theoretical capacities of AI blind us to present realities. Saying that current large language models understand is, as \citet{mcdermott_artificial_1976} described a while back, another case of a "wishful mnemonic."

\subsection{Pragmatic NLP}\label{sec:future}


Instead of denying or abandoning understanding, \textit{representation} advances a science. It allows us to answer: how does this system understand? How similar are the representations of these two systems? This is what \citet{harnad_symbol_1990} and \citet{santoro_symbolic_2022} call for. Indeed, \textit{as-representation} describes emerging trends in NLP which \ldots






\textbf{Probe model internals.} While most benchmark tasks focus on input-output reliability \citep{linzen_how_2020, zhang_understanding_2021}, investigating understanding will require functional analysis. \citet{buckner_understanding_2020} calls for us to determine a taxonomy of the non-robust features detected by deep nets.
\citet{beckers_approximate_2020} show how to compare causal models at multiple levels of granularity. \citet{geiger_causal_2021}, \citet{li_implicit_2021}, and \citet{lovering_unit_2022} extend this analysis with interventions to ask which representations (simpler models) approximate large language models.
\citet{olsson_-context_2022} find evidence for so-called induction heads in transformer models. \citet{johnston_abstract_2022} find abstract "representations" emergent from neural networks trained on similar tasks. \citet{merrill_effects_2021} investigate norm growth saturation in transformers as their inductive biases.

\textbf{Add more of human language} such as
intersubjective, multi-agent environments. \citet{noukhovitch_emergent_2021} find that partially-competitive agents learn to use symbols.
In a text game, \citet{hendrycks_what_2021} gauge moral valence. This is as \citet{firestone_performance_2020} describes, to use species-fair human-machine comparisons.




We might focus on human data constraints, such as the CHILDES database of language learning \citep{macwhinney_childes_2000, linzen_how_2020}.
\citet{hill_environmental_2020} show how the increased modality of data to a deep network may lead to generalizability. Contra strict composition, \citet{santoro_symbolic_2022} argue for the inclusion of "socio-cultural interactions." 
These are similar to calls for dynamic grounding \citep{chandu_grounding_2021} and "common sense" \citep{sap_commonsense_2020}.
For example, I would like models which not only resolve gaze \citep{koleva_impact_2015} but also deploy sharing gaze to sharing in other domains. 


On generalizability \citet{mitchell_artificial_2019} asks us to consider micro domains: "abc:abd; xxyyzz:?". \citet{sahin_puzzling_2020} propose a task with a small data-set from Rosetta Stone.
\citet{brachman_machines_2022} propose using only the kids version of Wikipedia,  KidzSearch, to reduce the number of entities on which to train. A language performant agent should be able to do well in this micro-domain alone.



\textbf{Measure what models can learn.} I would like more work on the tractability of meaning along the lines of \citet{merrill_provable_2021} on a language of assertions. How many different streams of data (or "world scopes" \citealt{bisk_experience_2020}) must we add to models to make them more reliable? What kind of scaling factors can we reasonably expect as we include more aspects of human language?



\section{Conclusion}

Present language models have limited access to meaning (\S \ref{sec:under-spec}) and  scale alone may not be sufficient to achieve human-level understanding (\S \ref{sec:scaling}), at least until we can guarantee similar representations or inductive biases (\S \ref{sec:views}).

Inference tasks 
reduce the space of possible entities implied by conventional usage.
Still, we must be assured of what they represent in order to guarantee their reliability.
\textit{Understanding-as-mapping} would deny the distributional compositionality of our languages and of our minds.
\textit{Understanding-as-reliability} would claim our machines "get" meaning so long as we focus only on temporally limited usage.
\textit{Understanding-as-representation} would focus on accurate, measurable, and tractable meaning.







\section*{Limitations}

This is only one perspective of many possible on the nature of understanding. Given the short form of this presentation I am not able to do justice to the diverse fields--from philosophy of science to linguistics to brain science--which I reference. I may therefore insufficiently explain the relevant literature to my target readers in NLP. While I have attempted to present a well-informed prior on the future direction of AI, my perspective is nonetheless uncertain; I may ultimately be incorrect.

\section*{Acknowledgements}

Julian Michael, Ari Holtzman, Dallas Card, Johan Michalove, and Blaise Agüera y Arcas provided the invaluable conversations, questions, and challenges that have made the ideas of this paper what they are. The participants of Melanie Mitchell's "Embodied, Situated, and Grounded Intelligence" workshop at the Santa Fe institute helped me to realize the need for work such as this. I would also like to thank suggestions of my three anonymous reviewers.

\bibliography{custom}

\end{document}